\documentclass[runningheads]{llncs}
\usepackage{graphicx}
\usepackage{subfigure}
\usepackage{float}
\usepackage{graphicx,epsfig}
\usepackage{algorithm}
\usepackage{algorithmic}
\usepackage{amsmath}
\usepackage{color}
\usepackage{makecell}
\usepackage{bm}
\usepackage{multirow}
\usepackage{cite}
\usepackage{bbding}
\usepackage{verbatim}
\usepackage{makecell}

\begin{document}
\title{Leveraging Structure Knowledge and Deep Models for the Detection of Abnormal Handwritten Text\thanks{Supported by the National Natural Science Foundation of China under Grant No. 62106031 and the volunteers for the creation of the datasets. The author would like to thank professor Jun Du of University of Science and Technology of China and the anonymous reviewers for their constructive suggestions and comments.}}

\titlerunning{The Detection of Abnormal Handwritten Text}
% If the paper title is too long for the running head, you can set
% an abbreviated paper title here
%
\author{Zi-Rui Wang}
%
%\authorrunning{Zi-Rui Wang.}
% First names are abbreviated in the running head.
% If there are more than two authors, 'et al.' is used.
%
%\institute{Chongqing University of Posts and Telecommunications, Chongqing, China

\institute{Chongqing University of Posts and Telecommunications, Chongqing, China
	\email{cs211@mail.ustc.edu.cn}\\
}

\maketitle              % typeset the header of the contribution
\begin{abstract}
Currently, the destruction of the sequence structure in handwritten text has become one of the main bottlenecks restricting the recognition task. The typical situations include additional specific markers (the text swapping modification) and the text overlap caused by character modifications like deletion, replacement, and insertion. In this paper, we propose a two-stage detection algorithm that combines structure knowledge and deep models for the above mentioned text. Firstly, different structure prototypes are roughly located from handwritten text images. Based on the detection results of the first stage, in the second stage, we adopt different strategies. Specifically, a shape regression network trained by a novel semi-supervised contrast training strategy is introduced and the positional relationship between the characters is fully employed.
%: for the text swapping modification, based on a shape regression network and the relationship between the special marker and characters, we can obtain a more accurate result. Specially, a novel semi-supervised contrast training strategy is proposed for the shape regression network; for the overlapping text, we further employ the positional relationship between the characters. 
Experiments on two handwritten text datasets show that the proposed method can greatly improve the detection performance. The new dataset is available at https://github.com/Wukong90.

\end{abstract}
\section{Introduction}
With the emergence of deep learning \cite{lecun2015deep} in recent years, intelligent handwritten document processing has also achieved a new milestone. For example, the dewarp technology was invented to generate flat text images by using deep neural networks to predict pixels or key points and representative text detection work \cite{zhong2017deeptext,jiang2018r} used the Faster R-CNN\cite{ren2015faster}/Mask R-CNN\cite{he2017mask} framework to obtain candidate boxes in the first step, and then make further predictions. Based on the extracted text lines, the text recognition task is very important in many common scenes. However, for some seriously degenerated images, the corresponding results are significantly decreased. 

\begin{figure}[t]
	\centering
	\includegraphics[width=4.6in,height=2.2in]{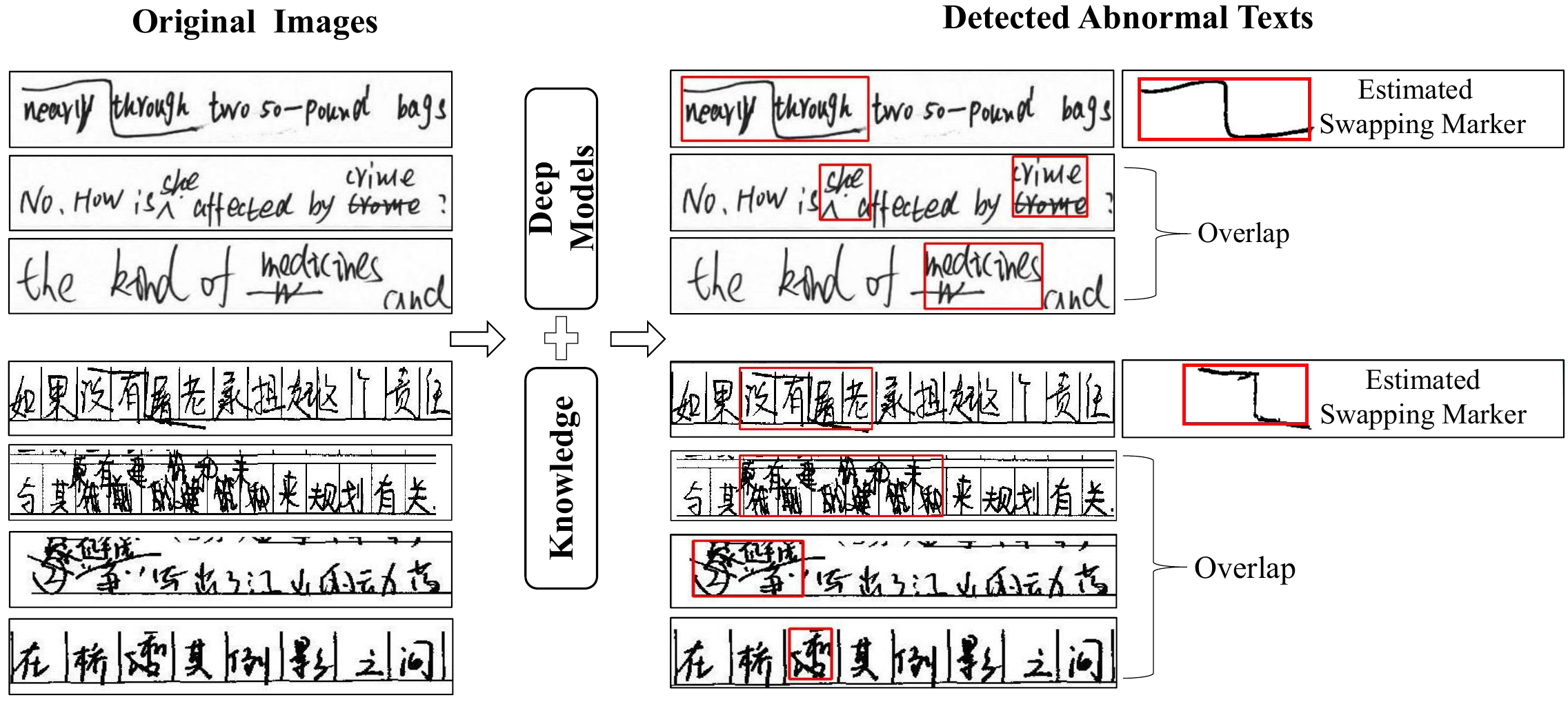}
	\caption{Some typical modifications are effectively detected by the proposed method.}
	\label{UnTe}
\end{figure}

The current challenges in the handwritten text recognition task are mainly caused by different modifications. The typical situations include additional markers and the text overlap caused by the text deletion, replacement, and insertion operations, which can produce irregular structures and disrupt the semantic order of the original text. As shown in Fig.~\ref{UnTe}, these abnormal parts consist of necessary text content that needs to be further processed and can not be simply removed. In order to effectively address the above problem, there exists the following challenges:

\begin{itemize}
	\item Compared with all text, the modified targets only occupy a small section. 
	\item The various modifications are highly similar. 
	\item The variations produced by free writing are infinite.
\end{itemize}

Different from Yan et al. \cite{yan2021recognizing} regarded the text recognition as a two-dimensional problem and tried to correctly recognize modified text with the structural attention network. In this paper, we directly focus on the detection of the text swapping modification and the text overlap. Compared with Hu et al. \cite{hu2021recyclenet} proposed a single ReycleNet to extract and reconstruct overlapped text instances, we propose a two-stage detection algorithm that explicitly combines structure knowledge and deep models to detect them effectively. Namely, the rough estimation of a specific shape (structure prototype) is obtained in the first stage. Based on the results of the first stage, different detection schemes are further adopted in the second stage according to the structure knowledge of different classes. In summary, the main contributions of this paper are as follows:

\begin{itemize}
	\item[(1)]For the abnormal handwritten text, we closely combine structure knowledge and deep models, and design a two-stage detection algorithm. The proposed algorithm greatly improves the detection performance.
	\item[(2)]According to the different shapes of the abnormal text, we define the corresponding structure prototypes.
	\item[(3)]In order to effectively train the deep networks, two intuitive data augmentation methods and the semi-supervised contrast training are proposed.
\end{itemize}

The remainder of this paper is organized as follows: Section \ref{Me} elaborates on the details of the proposed method. Section \ref{exp} reports the experimental results. Finally, we concludes the paper.

\section{Method}
\label{Me}
In this section, we elaborate on the details of the proposed two-stage detection algorithm. The algorithm highly depends on the structure knowledge of the abnormal text and the deep models.

\subsection{The first-stage detection based on the structure prototypes}

\begin{figure}
	\centering
	\includegraphics[width=4.1in,height=2.4in]{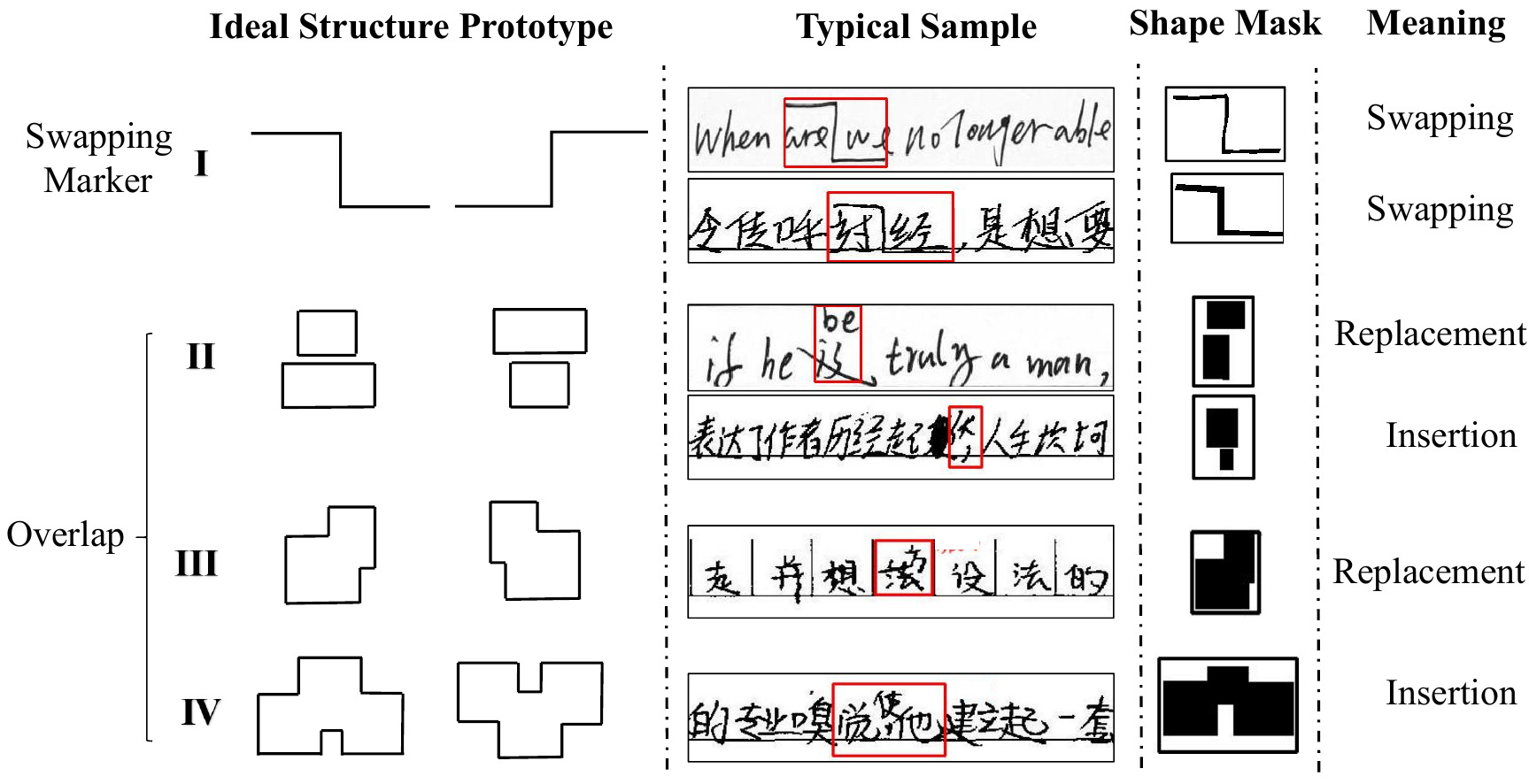}
	\caption{According to the various distinct shapes, we can define structure prototypes. The same overlap prototype may correspond to different modifications. In our experiments, the English dataset includes types I, II and the Chinese dataset includes types I-IV.}
	\label{Types}
\end{figure}

\begin{figure}
	\centering
	\includegraphics[width=3.1in,height=1.5in]{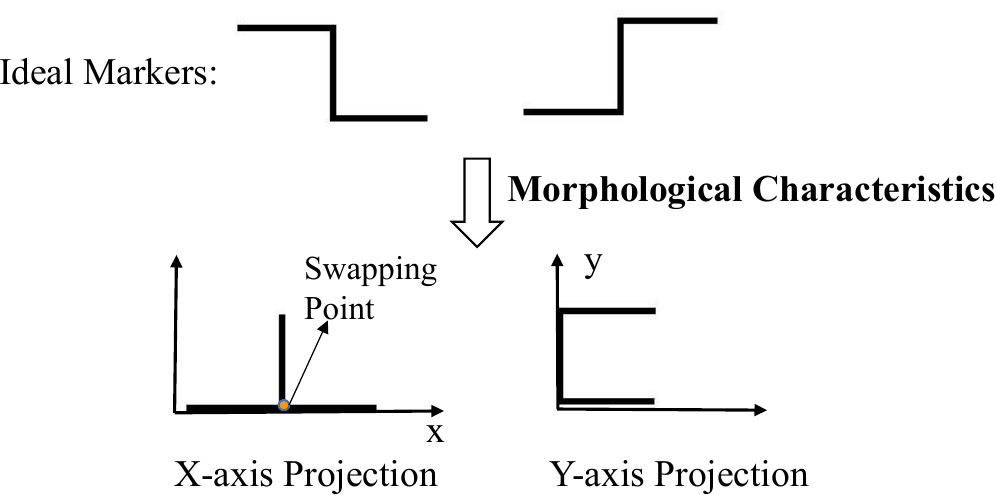}
	\caption{The morphological characteristics of the ideal markers.}
	\label{ideswi}
\end{figure}

The detection targets in the first stage are the abnormal structure prototypes that have obvious discriminative shapes. For text modifications with specific markers, the corresponding prototypes can be directly defined as these specific markers. As shown in Fig.~\ref{Types}, for an ideal swapping marker, in addition to its unique shape, its projection characteristics (Fig.~\ref{ideswi}) can be observed: the shape of the X-axis projection has one peak while the shape of the Y-axis projection has two peaks. This feature can provide the rules for accurate estimation of the marker mask. For text overlap, we determine three prototypes (Fig.~\ref{Types}) based on a large number of observations. Each type may correspond to different modifications (deletion, replacement or insertion). Their mirror symmetry are considered to belong to the same class, the left and right symmetry of prototypes I, III, and the up and down symmetry of prototypes II, IV. The combination of these simple overlap prototypes can form complex text structures. Please note that if the detection targets of the text overlap are set as their physical meanings, different classes may share many similar instances.

In this paper, we use the Mask-RCNN\cite{he2017mask} to detect the prototypes and denote the network as PDNN$(\bf{\Theta})$. In order to effectively train the network under the limited abnormal text samples, we propose two intuitive and practical data augmentation strategies, namely scale expansion and dynamic position change. The scale expansion is only designed for the swapping marker.

\subsubsection{Scale Expansion:}
For a standard swapping marker in Fig.~\ref{Syn_al}, considering a curve in practice occupies a certain area, there are two points on the most left edge of the curve in Fig.~\ref{Syn_al}-(1): ${\bf{p}}_1=(x_1,y_1)$, ${\bf{ p}}_1^{*}=(x_1^{*},y_2^{*})$. We suppose that the point ${\bf{p}}_{-1}=(x_{-1},y_{-1})$ is reached after a one-step extension, where $x_{-1}=x_1-\tau$. Meanwhile, the x-coordinate of the point ${\bf{p}}_{1}^{*}=(x_{1}^{*},y_{2}^{*})$ also moves the same distance: $x_{-1}^{*}=x_1^{*}-\tau$. The ordinates of the two extended points are randomly sampled from a small interval $(y_{1} - d, y_{1} + d) $, $(y_{1}^{*} - d, y_{1}^{*} + d)$:
\begin{eqnarray}
	\label{dyloc}
	{y_{-1}} = {y_{1}} + \varepsilon  {\rm{}}(\varepsilon {\rm{ }} \in [ - d,{\rm{ }}d])\\
	{y_{-1}^{*}} = {y_{1}^{*}} + \lambda  {\rm{}}(\lambda {\rm{ }} \in [ - d,{\rm{ }}d])
\end{eqnarray}
Similarly, the right-extension points ${\bf{q}}_{-1}$ and ${\bf{q}}_{-1}^{*}$ can be obtained by the same method. Then, we can fill the enlarged space. Furthermore, the above process is repeatedly conducted, and finally, a large number of training samples with different scales are obtained.

\subsubsection{Dynamic Location Change:}
\begin{figure}
	\centering
	\includegraphics[width=3.5in,height=1.6in]{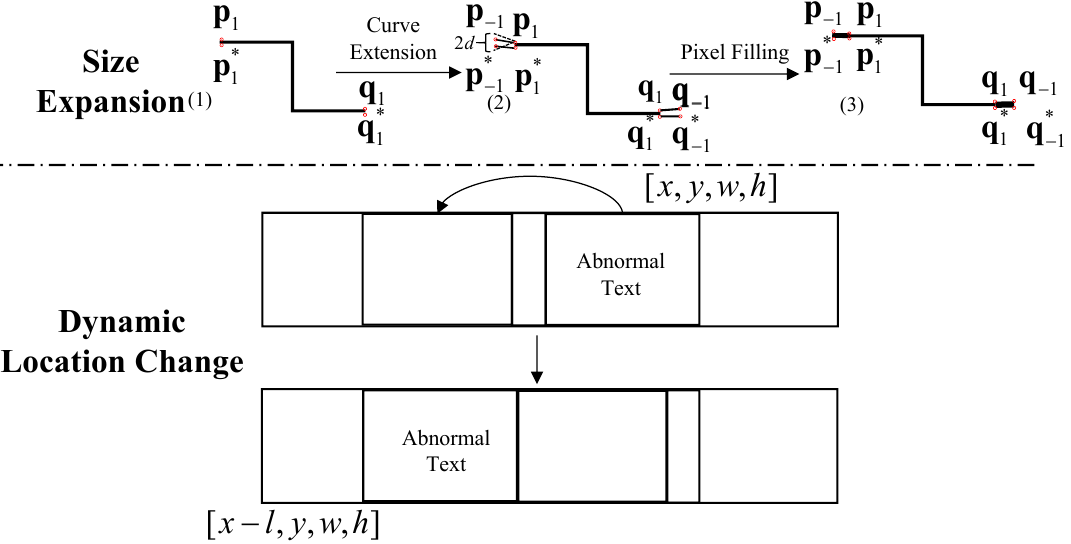}
	\caption{The proposed two data augmentation methods, the scale expansion, and the dynamic location change are illustrated.}
	\label{Syn_al}
\end{figure}

Considering the goal of the detection network is to achieve abnormal text detection, it has nothing to do with the information located outside a target box. Therefore, in order to further increase the diversity of samples, as shown in the lower part of Fig.~\ref{Syn_al}, we randomly move the position of the target box during the training stage. We use a rectangle box $[x,y,w,h]$ to represent the target box. $(x,y)$ is the coordinates of the upper left corner and $(w,h)$ is the corresponding width and height. During the training stage, the box is randomly shifted to the left or right by a distance of $l$. The figure shows that the target box moves to the left, and its surrounding content (light yellow and light green parts) also moves to the right by a distance of $l$.

\begin{figure*}
	\centering
	\includegraphics[width=5.3in,height=2.5in]{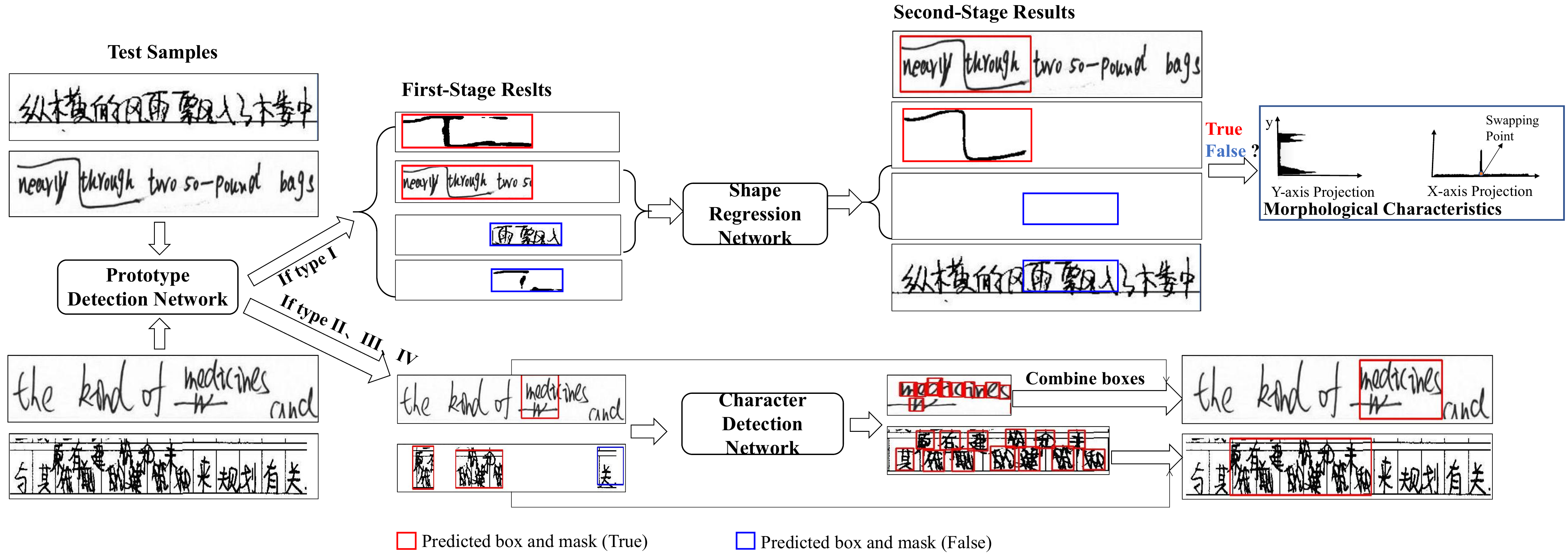}
	\caption{The pipeline of the proposed two-stage detection algorithm for the abnormal text.}
	\label{PipLin}
\end{figure*}

\subsection{The second-stage detection}
The detected results in the first stage often contain a large number of false positive boxes and the predicted boxes are not accurate enough. Therefore, based on the results of the first stage, we further conduct the second-stage detection. We design different strategies according to different structure knowledge.

\subsubsection{The precise detection for the text swapping}

The key goal in the second stage is to obtain accurate marker masks from the coarse deteciton results. By using the accurate marker masks, we can remove the false positive boxes and adjust the predicted boxes. As shown in Fig.~\ref{PipLin}, a shape regression network is employed to accurately estimate marker masks from the first-stage results. In order to effectively train the shape regression network, a semi-supervised contrast training strategy is proposed, namely, we randomly combine the positive samples and the negative samples from the outputs of the PDNN$(\bf{\Theta})$ to generate contrast images. After fedding the enhanced training set $S$ into the PDNN$(\bf{\Theta})$, the outputs belonging to the swapping modification constitute of a new set $D$:

\begin{eqnarray}
	\label{semicont}
	D=\rm{DPNN}(S|{\bf{\Theta}}) \quad ({\text{Predicted Labels}}(D)={\text{Type I}})
\end{eqnarray}

Then, we select the positive samples $D^{+}$ and the negative samples $D^{-}$. Furthermore, we randomly combine the positive samples $D^{+}$ and the negative samples $D^{-}$ to generate contrast training data $D^{*}$:

\begin{eqnarray}
	\label{randsamp}
	D^{*}= Sampling(D^{+}) \oplus Sampling(D^{-})
\end{eqnarray}

The contrast training set $D^{*}$ is used to train the shape regression neural network SRNN$(\bf{\Gamma})$. A typical U-Net is employed as the regression network. The training loss directly penalizes at each position $k$ of the input by using the cross-entropy:

\begin{eqnarray}
	&\hspace{-0.4cm}{L_{{\rm{SRNN}}}} =  - \sum\limits_k {[{l_{k}}\log {p_{k}} + (1 - {l_{k}})\log (1 - {p_{k}})]} 
\end{eqnarray}
$l_{k}=0/1$, and $p_{k}$ is the prediction probability at the position $k$. 

\begin{figure}
	\centering
	\includegraphics[width=4.1in,height=1.3in]{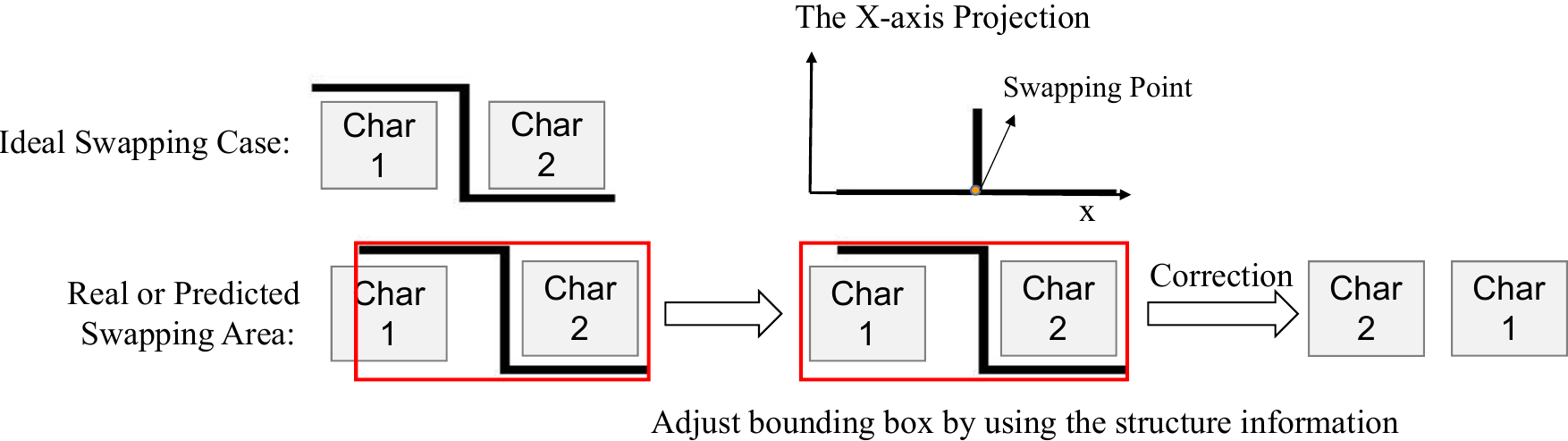}
	\caption{Employing the structure relationship between the marker mask and the characters to adjust the predicted box.}
	\label{strucswi}
\end{figure}

In the testing stage, we first analyze the marker masks from the SRNN$(\bf{\Gamma})$. As shown in Fig.~\ref{ideswi}, for a swapping marker, in addition to its unique shape, its projections have distinct characteristics. We simply judge whether a detected box really contains the modification target by the continuity of its projections. Meanwhile, we can adjust the predicted box according to the corresponding marker mask. Furthermore, a real modification area should be the structure in Fig.~\ref{strucswi}. Considering an incomplete modification, the bounding box may not cover all characters. We can further employ the positional relationship between the marker mask and the characters to adjust the predicted bounding box. In the experiments, a Faster-RCNN\cite{ren2015faster} is used as the character detection neural network (CDNN). Finally, by searching the maximum value on the horizontal projection, we can quickly find the swapping point and achieve text correction.

\subsubsection{The precise detection for the text overlap}
A prototype box $[x_{\rm{p}}^{'},y_{\rm{p}}^{'},w_{\rm{p}}^{'},h_{\rm{p}}^{'}]$ obtained in the first stage may be a false positive box, or it may contain a part of all stacked text. We can use the character detection network (CDNN) to detect characters within the prototype area and adjust the detected box based on the structure relationship of different characters. The detailed description is shown in Alg.\ref{alg:redet}, where $Q_{\rm{main}}$ and $Q_{\rm{ov}}$ respectively represent the character sequence of the main text line and the corresponding overlapping characters. Finally, according to the overlap of $Q_{\rm{main}}$ and $Q_{\rm{ov}}$ in the horizontal direction, we can obtain an accurate prediction box $[{x}^{*}, {y}^{*},{w}^{*},h^{*}]$. 

\begin{algorithm}
	\caption{The precise detection for the text overlap by combining results of different stages.}
	\label{alg:redet}
	\begin{algorithmic}[1]
		\REQUIRE~~\\
		A prototype detection box $[x_{\rm{p}}^{'},y_{\rm{p}}^{'},w_{\rm{p}}^{'},h_{\rm{p}}^{'}]$;\\
		The character boxes ${\rm{B}}_n([x_n^{'},y_n^{'},w_n^{'},h_n^{'}])$ $(n=1:N)$ within the prototype area;
		\STATE Calculate the geometric center for each character box:
		\begin{eqnarray}
			{{\rm{C}}_{{n_x}}} = x_n^{'} + \frac{{w_n^{'}}}{2}\\
			{{\rm{C}}_{{n_y}}} = y_n^{'} + \frac{{h_n^{'}}}{2}
		\end{eqnarray}
		\STATE Find the maximum and minimum values in the vertical direction of all geometric centers within the interval $(x_{\rm{p}}^{'}-\gamma,x_{\rm{p}}^{'}+\gamma)$:
		\begin{eqnarray}
			{y_{\rm{max}}} = {\rm{max}}({{\rm{C}}_{{n_y}}})\quad(x_{\rm{p}}-\gamma<={{\rm{C}}_{{n_x}}}<=x_{\rm{p}}+\gamma)\\
			{y_{\rm{min}}} = {\rm{min}}({{\rm{C}}_{{n_y}}})\quad(x_{\rm{p}}-\gamma<={{\rm{C}}_{{n_x}}}<=x_{\rm{p}}+\gamma)
		\end{eqnarray}
		\STATE If $({y_{\rm{max}}}-{y_{\rm{min}}}<=\alpha)$:\\
		\STATE \quad {\textbf{return}}\quad This is a false positive box.
		\STATE Else:\\
		\STATE\quad Build queues $Q_1$,$Q_2$;\\
		\STATE\quad Traverse all character boxes in turn:
		\STATE\quad\quad If $({y_{\rm{max}}}-{{\rm{C}}_{{n_y}}}<={y_{\rm{min}}}-{{\rm{C}}_{{n_y}}})$:\\
		\STATE\quad\quad\quad $Q_1$.append(${\rm{B}}_n$);\\
		\STATE\quad\quad Else:\\
		\STATE\quad\quad\quad $Q_2$.append(${\rm{B}}_n$);\\
		\STATE\quad If (the characters in the queues $Q_1$ and $Q_1$ have no overlap in the horizontal direction):\\
		\STATE \quad \quad {\textbf{return}}\quad This is a false positive box.
		\STATE\quad Else: \\
		\STATE\quad\quad If (len($Q_1$)$>$len($Q_2$)): \\
		\STATE\quad\quad\quad $Q_{\rm{main}}=Q_1$\\
		\STATE\quad\quad\quad In the queue $Q_2$, remove discontinuous character boxes (an interval greater than $\beta$);\\
		\STATE\quad\quad\quad $Q_{\rm{ov}}=Q_2$\\
		\STATE\quad\quad Else: \\
		\STATE\quad\quad\quad $Q_{\rm{main}}=Q_2$\\
		\STATE\quad\quad\quad In the queue $Q_1$, remove discontinuous character boxes (an interval greater than $\beta$);\\
		\STATE\quad\quad\quad $Q_{\rm{ov}}=Q_1$\\
		\STATE\quad According to the overlapped characters in the queues $Q_{\rm{main}}$ and $Q_{\rm{ov}}$, obtain the final prediction box $[{x}^{*},{y}^{*},{w}^{*},{h}^{*}]$.\\
		\STATE \quad  {\textbf{return}}\quad The final prediction box $[{x}^{*},{y}^{*},{w}^{*},h^{*}]$.
	\end{algorithmic}
\end{algorithm}

%\begin{figure}
%\centering
%\includegraphics[width=2.5in,height=1.2in]{ovstru.pdf}
%\caption{纠正模块中，对有交叠区域的字符进行分离，图中假设只有两个字符相互交叠。}
%\label{strucov}
%\end{figure}

\section{Experiments}
\label{exp}

\subsection{Datasets and evaluation metrics}

\begin{table}
	\centering
	\caption{The number of the abnormal text in the EHT and the SCUT-EPT datasets. In the SCUT-EPT test set, we remove some overlapping text (216 text lines) with obviously incomplete characters during the evaluation of the text overlap.}
	\scalebox{0.85}{
	\begin{tabular}{|c|cc|cc|}
		\hline
		\multirow{2}{*}{Type}      & \multicolumn{2}{c|}{EHT}                                                           & \multicolumn{2}{c|}{SCUT-EPT}                                                      \\ \cline{2-5} 
		& \multicolumn{1}{c|}{Training Set}  &Test Set & \multicolumn{1}{c|}{Training Set} & Test Set \\ \hline
		Text Line                  & \multicolumn{1}{c|}{209}                      & 188      & \multicolumn{1}{c|}{681}                     & 704      \\ \hline
		Text Swapping & \multicolumn{1}{c|}{61}                       & 52       & \multicolumn{1}{c|}{50}                       & 80       \\ \hline
		Text Overlap & \multicolumn{1}{c|}{179}                   & 147      & \multicolumn{1}{c|}{759}                    & 714       \\
		\hline
	\end{tabular}
}
	\label{Stanum}
\end{table}

We created an English handwritten text (EHT) data set that includes some abnormal text. According to the prepared content, the EHT samples were written by 9 adults. Each person was told to complete ten different pages in the specified blank lines. The content consists of well-known sayings, news reports, excerpts from famous books, etc. They were randomly collected from the Internet. The volunteers were told to randomly generate text swapping and overlap. Then, we segmented these page-level texts into lines. According to the differences of writers, all data were split into the training data and the test data. Before training, the text-line images were adjusted to 1185 x 99 size.

The Chinese handwritten text data set SCUT-EPT\cite{zhu2018scut} contains a large number of text modifications (text swapping, replacement, etc.), which is suitable for our research. It contains 4,250 categories of Chinese characters and symbols, totaling 1,267,161 characters, and it was split into 40,000 training text lines and 10,000 test text lines. These samples were written by 2,986 students, and the training and test sets do not include the same writers. This dataset only provides text lines and the corresponding annotations. We selected the abnormal text from the original images and supplemented the corresponding boxes and masks. Before training, all images were first resized to 1440 x 96.

Table~\ref{Stanum} lists the number of abnormal text in both datasets. During the training stage, we randomly used 70\% of all samples in the training set as training and 30\% of the data as validation. According to the performance indicators of the validation set, the optimal model was selected to predict the samples in the test set. On both datasets, we used the same networks. The Mask-RCNN and the Faster-RCNN with the same feature pyramid structure was used as the PDNN and the CDNN, respectively. The U-Net \cite{ronneberger2015u} was the SRNN. For the Mask-RCNN, the specific parameter configurations of the backbone network (ResNet18), the RPN, the ROIAlign and the classification branch, the detection box regression branch, and the mask branch were consistent with those provided by PyTorch \cite{paszke2019pytorch}. The detection indicators include precision, recall, and F1-score. In all experiments, the Intersection over Union (IoU) is set to 0.5 or 0.75.

%\begin{eqnarray}
%	\label{met1}
%	&&\hspace{-0.5cm}{\rm{Precision}} = \frac{{{\rm{True \; Positive %}}}}{{{\rm{True \; Positive}} + {\rm{False \;  Positive}}}} \\
%	\label{met2}
%	&&\hspace{-0.0cm}{\rm{Recall}} = \frac{{{\rm{True \; Positive %}}}}{{{\rm{Total\; Positive \; Samples }}}}\\
%	\label{met3}
%	&&\hspace{-0.4cm}{\rm{F1\text{-}score}} = \frac{{{\rm{2 *Precision * %Recall}}}}{{{\rm{Precision + Recall}}}}
%\end{eqnarray}
%{\color{red}{
		%进一步地，我们定义了第二类指标(公式（\ref{met4}）-(\ref{met6}))：完全正确纠正率（complete correction rate, CCR），部分正确矫正率（partial correction rate，PCR）和错误矫正率（error correction rate，ECR）。完全正确纠正意味着可以完整、准确地矫正修改的内容，即，修改标记完全覆盖修改的内容，并且由所提算法很好地预测了相应的边界框和标记掩膜。部分正确矫正意味着仅矫正了修改内容的部分，这可能是由于修改标记符的书写不完整，或者是算法预测结果不完整。错误矫正率是矫正内容包含了没有进行修改的文本占所有被矫正样品数量的比率。
		
		%\begin{eqnarray}
		%\label{met4}
		%&&{\rm{CCR}} = \frac{{{\rm{\text{完全正确纠正样本}}}}}{{{\rm{\text{所有测试集修改样本}}}}}\\
		%\label{met5}
		%&&{\rm{PCR}} = \frac{{{\rm{\text{部分正确纠正样本}}}}}{{{\rm{\text{所有测试集修改样本}}}}} \\
		%\label{met6}
		%&&{\rm{ECR}} = \frac{{{\rm{\text{错误纠正样本}}}}}{{{\rm{\text{所有被纠正的样本}}}}} 
		%\end{eqnarray}
		%}}
\subsection{The experimental results for the text swapping}

\begin{table}
	\centering
	\caption{The detection results of the text swapping on the data sets. The abbreviations SE, DLC, CT and SI represent size expansion, dynamic location change, contrast training and structure information, respectively. */* correspond to the different situations, namely, IoU=0.5/0.75.}
	\scalebox{0.85}{
		\begin{tabular}{|cc|c|c|c|}
			\hline
			\multicolumn{2}{|c|}{Method} & Metric (\%) & EHT & SCUT-EPT \\ \hline
			
			\multicolumn{1}{|c|}{\multirow{6}{*}{\begin{tabular}[c]{@{}c@{}}The first stage\end{tabular}}} & \multirow{3}{*}{Without SE and DLC} & Precision   & 47.2/35.8  &  11.8/4.8      \\
			\multicolumn{1}{|c|}{}                                                                &                            & Recall   & 96.2/73.1 &  53.8/21.3      \\
			\multicolumn{1}{|c|}{}                                                                &                            & F1-score & 63.3/48.1  &   19.4/7.8     \\ \cline{2-5} 
			\multicolumn{1}{|c|}{}                                                                & \multirow{3}{*}{SE and DLC}    & Precision  & 90.4/75.0 &    49.6/23.7    \\
			\multicolumn{1}{|c|}{}                                                                &                             & Recall   &  90.4/75.0  &     81.3/38.8    \\
			\multicolumn{1}{|c|}{}                                                                &                             & F1-score & 90.4/75.0  &   61.6/29.4       \\ \hline\hline
			
			\multicolumn{1}{|c|}{\multirow{9}{*}{\begin{tabular}[c]{@{}c@{}}The second stage\end{tabular}}} & \multirow{3}{*}{Without CT and SI}   & Precision   & 94.2/82.7 & 49.6/27.5         \\
			\multicolumn{1}{|c|}{}                                                               &                              & Recall   &  94.2/82.7 &   81.3/45.0      \\
			\multicolumn{1}{|c|}{}                                                               &                              & F1-score &  94.2/82.7 &   61.6/34.1     \\ \cline{2-5} 
			\multicolumn{1}{|c|}{}                                                                                     & \multirow{3}{*}{CT}            & Precision      & 100.0/94.0   &      83.6/47.3    \\
			\multicolumn{1}{|c|}{}                                                                                     &                                     & Recall                  & 96.2/90.4     &    57.5/32.5      \\
			\multicolumn{1}{|c|}{}                                                                                     &                                     & F1-score                  & 98.1/92.2           &     68.1/38.5     \\ \cline{2-5} 
			
			\multicolumn{1}{|c|}{}                                                                                     & \multirow{3}{*}{CT and SI}            & Precision       & \textbf{100.0/96.0}   &      \textbf{90.9/56.4}    \\
			\multicolumn{1}{|c|}{}                                                                                     &                                     & Recall      &  96.2/92.3   &   62.5/38.8       \\
			\multicolumn{1}{|c|}{}                                                                                     &                                     & F1-score             & \textbf{98.1/94.1}      &     \textbf{74.1/46.0}     \\ \hline
		\end{tabular}
		}
	\label{abla}
\end{table}

We conducted ablation experiments to verify the effectiveness of the proposed two-stage algorithm. Table~\ref{abla} lists the relevant results. In the scale expansion, the pixel value was extended from 2 to 150. We sampled every 5 pixels, and $d$ was set to 1 in Formula~\ref{dyloc}. 

\subsubsection{One-Stage vs. Two-Stage}
\begin{figure*}
	\centering
	\includegraphics[width=4.8in,height=2.5in]{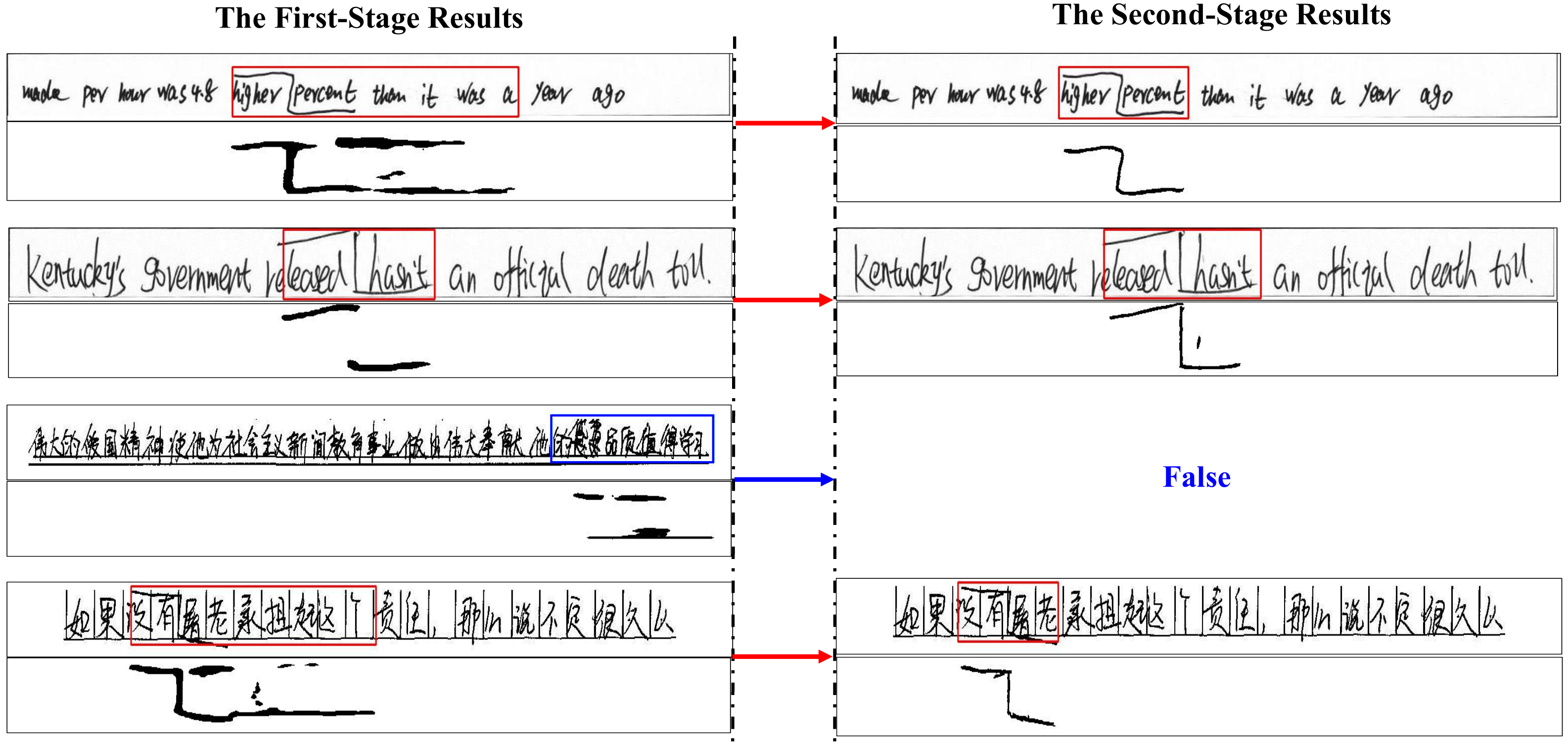}
	\caption{The comparisons of the one-stage results and the two-stage results. False means that the detected boxes from the one-stage network do not include any targets.}
	\label{OneVTwo}
\end{figure*}

Compared with the initial single-stage detection performance, the proposed algorithm greatly improve the detection precision and F1-score (IOU=0.5/0.75 in all experiments). On the English data set, the precision is increased from 47.2\%/35.8\% to 100.0\%/96.0\%, and the F1-score is increased from 63.3\%/48.1\% to 98.1\%/94.1\%; On the Chinese data set, the precision is improved from 11.8\%/4.8\% to 90.9\%/56.4\%, and the F1-score is increased from 19.4\%/7.8\% to 74.1\%/46.0\%. Fig.~\Ref{OneVTwo} shows the comparison between the detection results of the one-stage and the results of the two-stage. We can observe that the detected boxes from the PDNN containing non-target text can be effectively removed through the second stage. Besides, completely and accurately marker masks can be estimated after the second stage.

%\begin{figure}
%	\centering
%	\includegraphics[width=4.1in,height=2.3in]{syn_o.pdf}
%	\caption{The synthesized data with different sizes and locations. The %text images without red boxes are the original samples.}
%	\label{syn_o}
%\end{figure}

\begin{figure}
	\centering
	\includegraphics[width=4.2in,height=1.5in]{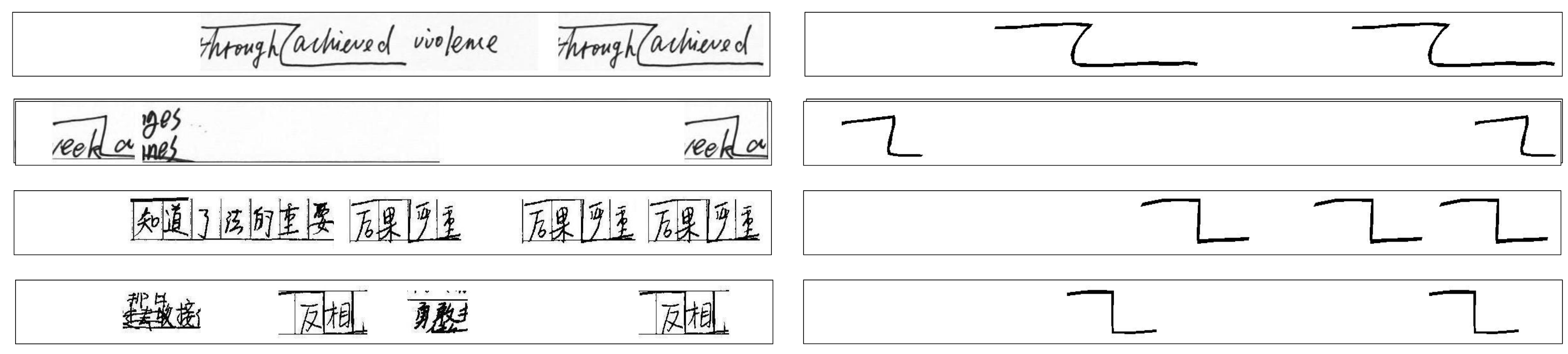}
	\caption{The synthesized contrast training data by using the predicted results of the PDNN.}
	\label{syn_t}
\end{figure}

\subsubsection{With/Without the SE and the DLC}As shown in Table~\ref{abla}, the data augmentation can significantly improve the precision and F1-score. The precision on the English data set is increased from 47.2\%/35.8\% to 90.4\%/75.0\%, and the F1-score is increased from 63.3\%/48.1\% to 90.4\%/75.0\%. The precision on the Chinese data set is increased from 11.8\%/4.8\% to 49.6\%/23.7\%, and the F1-score is increased from 19.4\%/7.8\% to 61.6\%/29.4\%. 
%Fig.~\Ref{syn_o} shows the synthesized data with different sizes and locations.

\begin{figure}
	\centering
	\includegraphics[width=4.0in,height=1.2in]{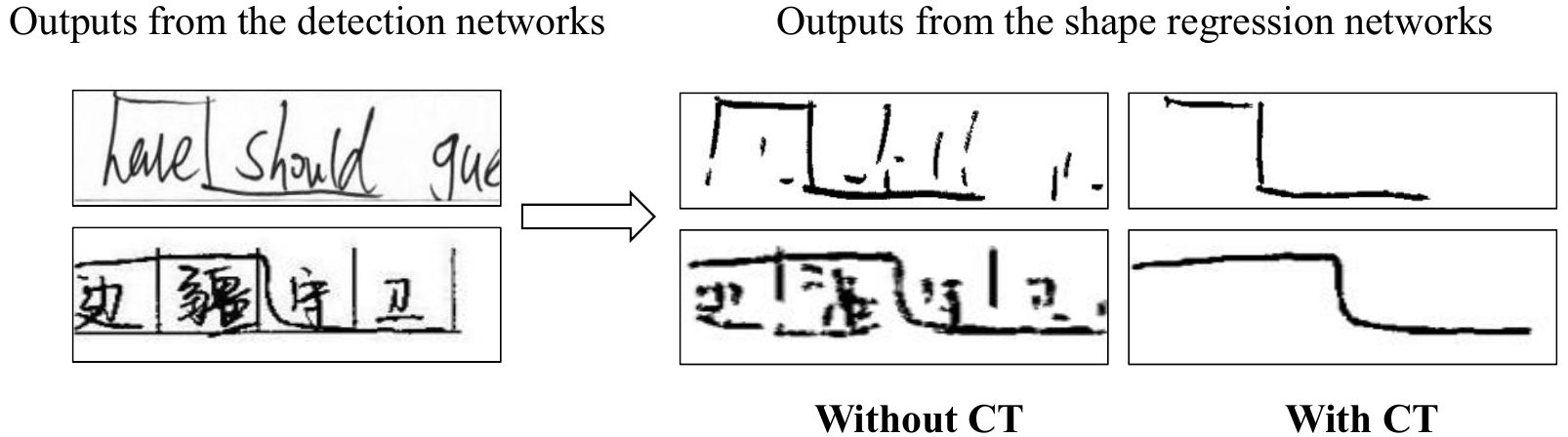}
	\caption{The outputs of the shape regression networks with/without CT.}
	\label{Outputs}
\end{figure}

\subsubsection{With/Without the CT}As shown in Fig.~\ref{Outputs}, the CT is important for the training of the SRNN. The training loss without the CT was always high and it was difficult to converge to a satisfactory value. A well trained SRNN can generate accurate marker masks, which enables us to conduct the subsequent processes, namely, the detected box judgment, adjustment, and text correction.

\subsubsection{With/Without the Structure Information}
\begin{figure*}[htbp]
	\centering
	\includegraphics[width=4.8in,height=1.6in]{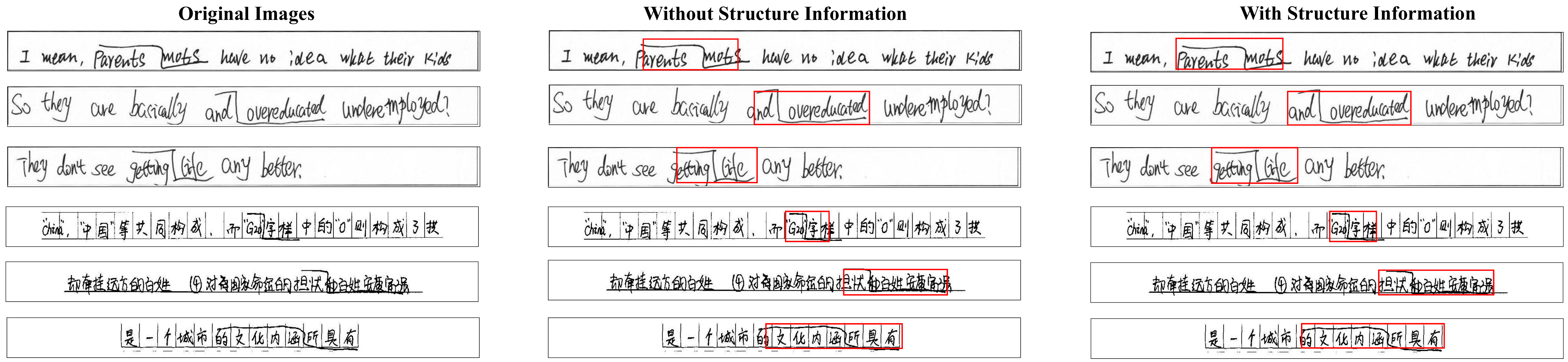}
	\caption{By using the structure information, many incomplete predictions can be improved.}
	\label{PCTtCCR}
\end{figure*}

After using the structure information between the characters and the marker (Fig.~\ref{strucswi}) to make a final adjustment, the main detection indicators can be improved. On the EHT data set, the precision and the F1-score at an IOU of 0.7 are increased from 94.0\%, 92.2\% to 96.0\%, 94.1\%, respectively. On the SCUT-EPT test set, all indicators are significantly improved, and the precision and the F1-score are increased from 83.6\%/47.3\%, 68.1\%/38.5\% to 90.9\%/56.4\%, 74.1\%/46.0\%, respectively. As shown in Fig.\ref{PCTtCCR}, after using the structure information, a considerable number of incomplete predictions are corrected. There are two main reasons for incomplete predictions: 1) the written modification markers do not cover all characters; 2) the algorithm do not completely predict the entire modification content and marker masks. 

\subsection{The experimental results for the text overlap}
Table~\ref{ovre} lists the relevant detection results of the text overlap.

\subsubsection{One-Stage vs. Two-Stage}
\begin{figure*}
	\centering
	\includegraphics[width=4.8in,height=3.3in]{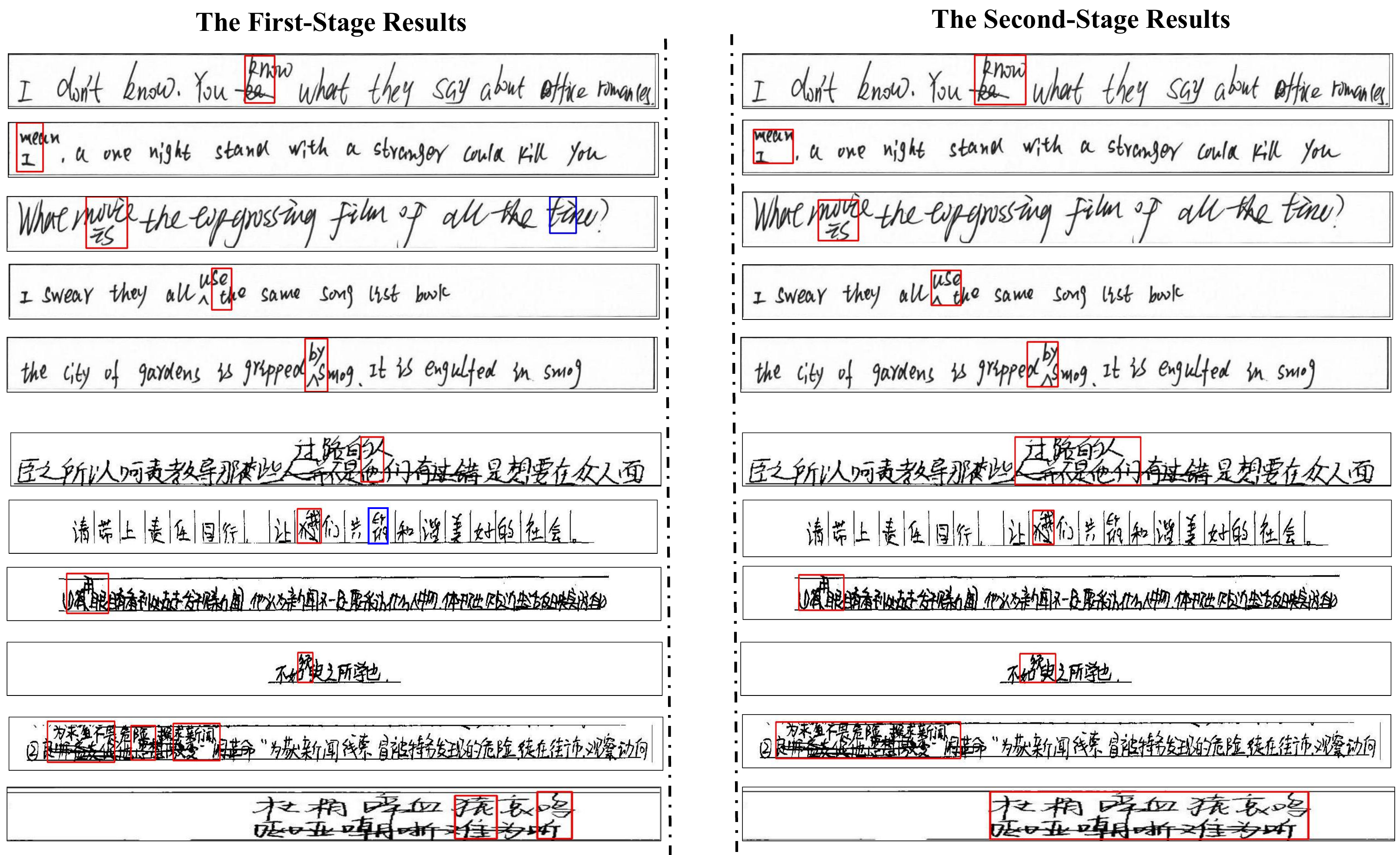}
	\caption{By combining the results of different stages, the false positive boxes in the first stage can be effectively removed and the complete predictions can be obtained.}
	\label{OneVTwo_ov}
\end{figure*}

Compared with the initial detection performance, the proposed algorithm greatly improve the detection precision and F1-score (IOU=0.5/0.75 in all experiments). On the English data set, the precision is increased from 62.5\%/26.1\% to 93.1\%/76.2\%, and the F1-score is increased from 69.5\%/29.0\% to 87.4\%/71.5\%; on the Chinese data set, the precision is improved from 27.7\%/12.3\% to 76.2\%/53.0\%, and the F1-score is increased from 35.9\%/16.0\% to 76.7\%/53.3\%. Fig.~\Ref{OneVTwo_ov} shows the comparisons of the first-stage resutls and the second-stage results. The false positive boxes in the first stage can be effectively removed and the complete predictions can be obtained by combining the two-stage results.

\begin{table}
	\centering
	\caption{The detection results of the text overlap on the data sets. */* correspond to the different situations, namely, IoU=0.5/0.75.}
	\scalebox{0.85}{
	\begin{tabular}{|cc|c|c|c|}
		\hline
		\multicolumn{2}{|c|}{Method} & {Metric (\%)} & EHT & SCUT-EPT \\ \cline{4-5} \hline
		
		\multicolumn{1}{|c|}{\multirow{6}{*}{\begin{tabular}[c]{@{}c@{}}One Stage\end{tabular}}} & \multirow{3}{*}{Without DLC} & Precision     & 62.5/26.1  &  27.7/12.3      \\
		\multicolumn{1}{|c|}{}                                                                &                         & Recall   &   78.2/32.7  &  51.0/22.7      \\
		\multicolumn{1}{|c|}{}                                                                &                         & F1-score  & 69.5/29.0   &   35.9/16.0       \\ \cline{2-5} 
		\multicolumn{1}{|c|}{}                                                               & \multirow{3}{*}{DLC}    & Precision      & 86.9/39.3 &     32.7/15.8    \\
		\multicolumn{1}{|c|}{}                                                                                     &    & Recall   &  85.7/38.8   &    58.4/28.3    \\
		\multicolumn{1}{|c|}{}                                                                                     &     & F1-score   &  86.3/39.0   &   41.9/20.3       \\ \hline\hline
		
		\multicolumn{1}{|c|}{\multirow{3}{*}{\begin{tabular}[c]{@{}c@{}}Two Stage\end{tabular}}} & \multirow{3}{*}{SI}         & Precision   &     \textbf{93.1/76.2} & \textbf{76.2/53.0}         \\
		\multicolumn{1}{|c|}{}                                                                                     &       & Recall     &  82.3/67.3  &   77.2/53.6      \\
		\multicolumn{1}{|c|}{}                                                                                     &       & F1-score  & \textbf{87.4/71.5}    &     \textbf{76.7/53.3}     \\ \cline{2-5} 
		\hline
	\end{tabular}
}
	\label{ovre}
\end{table}

\subsubsection{With/Without the DLC}On both data sets, the DLC can significantly improve the precision and F1-score (Table \ref{ovre}). The precision on the English data set is increased from 62.5\%/26.1\% to 86.9\%/39.3\%, and the F1-score is increased from 69.5\%/29.0\% to 86.3\%/39.0\%. The precision on the Chinese data set is increased from 27.7\%/12.3\% to 32.7\%/15.8\%, and the F1-score is increased from 35.9\%/16.0\% to 41.9\%/20.3\%.

\subsubsection{The second-stage results with the structure information}After using the structure information to make a further prediction, the detection performance can obtain an impressive gain. On the EHT data set, the precision is increased from 86.9\%/39.3\% to 93.1\%/76.2\%, and the F1-score is changed from 86.3\%/39.0\% to 87.4\%/71.5\%. On the SCUT-EPT test set, the precision is increased from 32.7\%/15.8\% to 76.2\%/53.0\%, and the F1-score is increased from 41.9\%/20.3\% to 76.7\%/53.3\%. As shown in Fig.~\Ref{OneVTwo_ov}, with the structure information, the false positive boxes are effectively removed and the complete predictions can be obtained. Therefore, the detection performance can be greatly improved.  

%Compared with the English language, the Chinese language is derived from one kind of hieroglyphs and it includes many different characters. In addition, the SCUT-EPT itself is a difficult data set, including scratchy handwriting and a large number of modifications. Therefore, the detection performance on the SCUT-EPT data set is lower than the performance on the EHT data set.

\subsection{The impact of the abnormal text on the recognition task}
Finally, in order to illustrate the importance of the effective detection of the abnormal text, we tentatively use a conventional CTC-based deep neural network to recognize the detected text swapping modification, and its training set consists of all the training images in three publicly available Chinese handwritten text datasets (CASIA\cite{liu2011casia}, SCUT-HCCDoc\cite{zhang2020scut} and SCUT-EPT\cite{zhu2018scut}). However, the conventional recognition model is basically unable to realize the change of the text order, so the accuracy rate (AR) is very low (8.2\%). After correcting the detected swapping text, the AR can achieve 69.2\%.

\section{Conclusion}
\label{Con}
In this paper, we conduct a comprehensive study on the detection of the text swapping modification and the text overlap. To detect them effectively, we propose a two-stage detection algorithm that combines structure knowledge and deep models. Experimental results demonstrate significant improvements.

%
% the environments 'definition', 'lemma', 'proposition', 'corollary',
% 'remark', and 'example' are defined in the LLNCS documentclass as well.
%
%
% ---- Bibliography ----
%
% BibTeX users should specify bibliography style 'splncs04'.
% References will then be sorted and formatted in the correct style.
%
% \bibliographystyle{splncs04}
% \bibliography{mybibliography}
%

\bibliographystyle{unsrt}
\bibliography{ref.bib}

\begin{thebibliography}{10}

\bibitem{lecun2015deep}
Yann LeCun, Yoshua Bengio, and Geoffrey Hinton.
\newblock Deep learning.
\newblock {\em nature}, 521(7553):436--444, 2015.

\bibitem{graves2006connectionist}
Alex Graves, Santiago Fern{\'a}ndez, Faustino Gomez, and J{\"u}rgen
  Schmidhuber.
\newblock Connectionist temporal classification: labelling unsegmented sequence
  data with recurrent neural networks.
\newblock In {\em Proceedings of the 23rd international conference on Machine
  learning}, pages 369--376, 2006.

\bibitem{messina2015segmentation}
Ronaldo Messina and Jerome Louradour.
\newblock Segmentation-free handwritten chinese text recognition with lstm-rnn.
\newblock In {\em 2015 13th International conference on document analysis and
  recognition (icdar)}, pages 171--175. IEEE, 2015.

\bibitem{shi2016end}
Baoguang Shi, Xiang Bai, and Cong Yao.
\newblock An end-to-end trainable neural network for image-based sequence
  recognition and its application to scene text recognition.
\newblock {\em IEEE transactions on pattern analysis and machine intelligence},
  39(11):2298--2304, 2016.

\bibitem{shi2018aster}
Baoguang Shi, Mingkun Yang, Xinggang Wang, Pengyuan Lyu, Cong Yao, and Xiang
  Bai.
\newblock Aster: An attentional scene text recognizer with flexible
  rectification.
\newblock {\em IEEE transactions on pattern analysis and machine intelligence},
  41(9):2035--2048, 2018.

\bibitem{wang2020writer}
Zi-Rui Wang, Jun Du, and Jia-Ming Wang.
\newblock Writer-aware cnn for parsimonious hmm-based offline handwritten
  chinese text recognition.
\newblock {\em Pattern Recognition}, 100:107102, 2020.

\bibitem{tian2020deep}
Chunwei Tian, Lunke Fei, Wenxian Zheng, Yong Xu, Wangmeng Zuo, and Chia-Wen
  Lin.
\newblock Deep learning on image denoising: An overview.
\newblock {\em Neural Networks}, 131:251--275, 2020.

\bibitem{loizou2007speech}
Philipos~C Loizou.
\newblock {\em Speech enhancement: theory and practice}.
\newblock CRC press, 2007.

\bibitem{zhang2017beyond}
Kai Zhang, Wangmeng Zuo, Yunjin Chen, Deyu Meng, and Lei Zhang.
\newblock Beyond a gaussian denoiser: Residual learning of deep cnn for image
  denoising.
\newblock {\em IEEE transactions on image processing}, 26(7):3142--3155, 2017.

\bibitem{xu2014regression}
Yong Xu, Jun Du, Li-Rong Dai, and Chin-Hui Lee.
\newblock A regression approach to speech enhancement based on deep neural
  networks.
\newblock {\em IEEE/ACM Transactions on Audio, Speech, and Language
  Processing}, 23(1):7--19, 2014.

\bibitem{zhu2018scut}
Yuanzhi Zhu, Zecheng Xie, Lianwen Jin, Xiaoxue Chen, Yaoxiong Huang, and Ming
  Zhang.
\newblock Scut-ept: New dataset and benchmark for offline chinese text
  recognition in examination paper.
\newblock {\em IEEE Access}, 7:370--382, 2018.

\bibitem{liu2007normalization}
Cheng-Lin Liu.
\newblock Normalization-cooperated gradient feature extraction for handwritten
  character recognition.
\newblock {\em IEEE transactions on Pattern Analysis and machine intelligence},
  29(8):1465--1469, 2007.

\bibitem{chin1997skew}
Wesley Chin, Alam Harvey, and Andrew Jennings.
\newblock Skew detection in handwritten scripts.
\newblock In {\em TENCON'97 Brisbane-Australia. Proceedings of IEEE TENCON'97.
  IEEE Region 10 Annual Conference. Speech and Image Technologies for Computing
  and Telecommunications (Cat. No. 97CH36162)}, volume~1, pages 319--322. IEEE,
  1997.

\bibitem{hull1998document}
Jonathan~J Hull.
\newblock Document image skew detection: Survey and annotated bibliography.
\newblock In {\em Document Analysis Systems II}, pages 40--64. World
  Scientific, 1998.

\bibitem{su2007skew}
T-H Su, T-W Zhang, H-J Huang, and Yu~Zhou.
\newblock Skew detection for chinese handwriting by horizontal stroke
  histogram.
\newblock In {\em Ninth international conference on document analysis and
  recognition (ICDAR 2007)}, volume~2, pages 899--903. IEEE, 2007.

\bibitem{yin2009handwritten}
Fei Yin and Cheng-Lin Liu.
\newblock Handwritten chinese text line segmentation by clustering with
  distance metric learning.
\newblock {\em Pattern Recognition}, 42(12):3146--3157, 2009.

\bibitem{zhong2017deeptext}
Zhuoyao Zhong, Lianwen Jin, and Shuangping Huang.
\newblock Deeptext: A new approach for text proposal generation and text
  detection in natural images.
\newblock In {\em 2017 IEEE international conference on acoustics, speech and
  signal processing (ICASSP)}, pages 1208--1212. IEEE, 2017.

\bibitem{wang2018comprehensive}
Zi-Rui Wang, Jun Du, Wen-Chao Wang, Jian-Fang Zhai, and Jin-Shui Hu.
\newblock A comprehensive study of hybrid neural network hidden markov model
  for offline handwritten chinese text recognition.
\newblock {\em International Journal on Document Analysis and Recognition
  (IJDAR)}, 21(4):241--251, 2018.

\bibitem{das2019dewarpnet}
Sagnik Das, Ke~Ma, Zhixin Shu, Dimitris Samaras, and Roy Shilkrot.
\newblock Dewarpnet: Single-image document unwarping with stacked 3d and 2d
  regression networks.
\newblock In {\em Proceedings of the IEEE/CVF International Conference on
  Computer Vision}, pages 131--140, 2019.

\bibitem{xie2021document}
Guo-Wang Xie, Fei Yin, Xu-Yao Zhang, and Cheng-Lin Liu.
\newblock Document dewarping with control points.
\newblock In {\em International Conference on Document Analysis and
  Recognition}, pages 466--480. Springer, 2021.

\bibitem{hu2021recyclenet}
Yiqing Hu, Yan Zheng, Xinghua Jiang, Hao Liu, Deqiang Jiang, Yinsong Liu,
  Bo~Ren, and Rongrong Ji.
\newblock Recyclenet: An overlapped text instance recovery approach.
\newblock In {\em Proceedings of the 29th ACM International Conference on
  Multimedia}, pages 1102--1110, 2021.

\bibitem{yan2021recognizing}
Shi Yan, Jin-Wen Wu, Fei Yin, and Cheng-Lin Liu.
\newblock Recognizing handwritten chinese texts with insertion and swapping
  using a structural attention network.
\newblock In {\em International Conference on Document Analysis and
  Recognition}, pages 557--571. Springer, 2021.

\bibitem{peng2022pagenet}
Dezhi Peng, Lianwen Jin, Yuliang Liu, Canjie Luo, and Songxuan Lai.
\newblock Pagenet: Towards end-to-end weakly supervised page-level handwritten
  chinese text recognition.
\newblock {\em International Journal of Computer Vision}, 130(11):2623--2645,
  2022.

\bibitem{bastien2010deep}
Fr{\'e}d{\'e}ric Bastien, Yoshua Bengio, Arnaud Bergeron, Nicolas
  Boulanger-Lewandowski, Thomas Breuel, Youssouf Chherawala, Moustapha Cisse,
  Myriam C{\^o}t{\'e}, Dumitru Erhan, Jeremy Eustache, et~al.
\newblock Deep self-taught learning for handwritten character recognition.
\newblock {\em arXiv preprint arXiv:1009.3589}, 2010.

\bibitem{haines2016my}
Tom~SF Haines, Oisin Mac~Aodha, and Gabriel~J Brostow.
\newblock My text in your handwriting.
\newblock {\em ACM Transactions on Graphics (TOG)}, 35(3):1--18, 2016.

\bibitem{Wigington2017data}
Curtis Wigington, Seth Stewart, Brian Davis, Bill Barrett, Brian Price, and
  Scott Cohen.
\newblock Data augmentation for recognition of handwritten words and lines
  using a cnn-lstm network.
\newblock In {\em 2017 14th IAPR international conference on document analysis
  and recognition (ICDAR)}, volume~1, pages 639--645. IEEE, 2017.

\bibitem{luo2020learn}
Canjie Luo, Yuanzhi Zhu, Lianwen Jin, and Yongpan Wang.
\newblock Learn to augment: Joint data augmentation and network optimization
  for text recognition.
\newblock In {\em Proceedings of the IEEE/CVF Conference on Computer Vision and
  Pattern Recognition}, pages 13746--13755, 2020.

\bibitem{goodfellow2014generative}
Ian Goodfellow, Jean Pouget-Abadie, Mehdi Mirza, Bing Xu, David Warde-Farley,
  Sherjil Ozair, Aaron Courville, and Yoshua Bengio.
\newblock Generative adversarial nets.
\newblock {\em Advances in neural information processing systems}, 27, 2014.

\bibitem{bhunia2019handwriting}
Ayan~Kumar Bhunia, Abhirup Das, Ankan~Kumar Bhunia, Perla Sai~Raj Kishore, and
  Partha~Pratim Roy.
\newblock Handwriting recognition in low-resource scripts using adversarial
  learning.
\newblock In {\em Proceedings of the IEEE/CVF Conference on Computer Vision and
  Pattern Recognition}, pages 4767--4776, 2019.

\bibitem{kang2020ganwriting}
Lei Kang, Pau Riba, Yaxing Wang, Mar{\c{c}}al Rusinol, Alicia Forn{\'e}s, and
  Mauricio Villegas.
\newblock Ganwriting: content-conditioned generation of styled handwritten word
  images.
\newblock In {\em European Conference on Computer Vision}, pages 273--289.
  Springer, 2020.

\bibitem{fogel2020scrabblegan}
Sharon Fogel, Hadar Averbuch-Elor, Sarel Cohen, Shai Mazor, and Roee Litman.
\newblock Scrabblegan: Semi-supervised varying length handwritten text
  generation.
\newblock In {\em Proceedings of the IEEE/CVF conference on computer vision and
  pattern recognition}, pages 4324--4333, 2020.

\bibitem{alonso2019adversarial}
Eloi Alonso, Bastien Moysset, and Ronaldo Messina.
\newblock Adversarial generation of handwritten text images conditioned on
  sequences.
\newblock In {\em 2019 international conference on document analysis and
  recognition (ICDAR)}, pages 481--486. IEEE, 2019.

\bibitem{luo2022slogan}
Canjie Luo, Yuanzhi Zhu, Lianwen Jin, Zhe Li, and Dezhi Peng.
\newblock Slogan: Handwriting style synthesis for arbitrary-length and
  out-of-vocabulary text.
\newblock {\em IEEE Transactions on Neural Networks and Learning Systems},
  2022.

\bibitem{redmon2016you}
Joseph Redmon, Santosh Divvala, Ross Girshick, and Ali Farhadi.
\newblock You only look once: Unified, real-time object detection.
\newblock In {\em Proceedings of the IEEE conference on computer vision and
  pattern recognition}, pages 779--788, 2016.

\bibitem{gupta2016synthetic}
Ankush Gupta, Andrea Vedaldi, and Andrew Zisserman.
\newblock Synthetic data for text localisation in natural images.
\newblock In {\em Proceedings of the IEEE conference on computer vision and
  pattern recognition}, pages 2315--2324, 2016.

\bibitem{liao2017textboxes}
Minghui Liao, Baoguang Shi, Xiang Bai, Xinggang Wang, and Wenyu Liu.
\newblock Textboxes: A fast text detector with a single deep neural network.
\newblock In {\em Thirty-first AAAI conference on artificial intelligence},
  2017.

\bibitem{liao2018textboxes++}
Minghui Liao, Baoguang Shi, and Xiang Bai.
\newblock Textboxes++: A single-shot oriented scene text detector.
\newblock {\em IEEE transactions on image processing}, 27(8):3676--3690, 2018.

\bibitem{liu2016ssd}
Wei Liu, Dragomir Anguelov, Dumitru Erhan, Christian Szegedy, Scott Reed,
  Cheng-Yang Fu, and Alexander~C Berg.
\newblock Ssd: Single shot multibox detector.
\newblock In {\em European conference on computer vision}, pages 21--37.
  Springer, 2016.

\bibitem{liao2018rotation}
Minghui Liao, Zhen Zhu, Baoguang Shi, Gui-song Xia, and Xiang Bai.
\newblock Rotation-sensitive regression for oriented scene text detection.
\newblock In {\em Proceedings of the IEEE conference on computer vision and
  pattern recognition}, pages 5909--5918, 2018.

\bibitem{zhou2017east}
Xinyu Zhou, Cong Yao, He~Wen, Yuzhi Wang, Shuchang Zhou, Weiran He, and Jiajun
  Liang.
\newblock East: an efficient and accurate scene text detector.
\newblock In {\em Proceedings of the IEEE conference on Computer Vision and
  Pattern Recognition}, pages 5551--5560, 2017.

\bibitem{he2017deep}
Wenhao He, Xu-Yao Zhang, Fei Yin, and Cheng-Lin Liu.
\newblock Deep direct regression for multi-oriented scene text detection.
\newblock In {\em Proceedings of the IEEE International Conference on Computer
  Vision}, pages 745--753, 2017.

\bibitem{jiang2018r}
Yingying Jiang, Xiangyu Zhu, Xiaobing Wang, Shuli Yang, Wei Li, Hua Wang, Pei
  Fu, and Zhenbo Luo.
\newblock R 2 cnn: Rotational region cnn for arbitrarily-oriented scene text
  detection.
\newblock In {\em 2018 24th International Conference on Pattern Recognition
  (ICPR)}, pages 3610--3615. IEEE, 2018.

\bibitem{ren2015faster}
Shaoqing Ren, Kaiming He, Ross Girshick, and Jian Sun.
\newblock Faster r-cnn: Towards real-time object detection with region proposal
  networks.
\newblock {\em Advances in neural information processing systems}, 28, 2015.

\bibitem{he2017mask}
Kaiming He, Georgia Gkioxari, Piotr Doll{\'a}r, and Ross Girshick.
\newblock Mask r-cnn.
\newblock In {\em Proceedings of the IEEE international conference on computer
  vision}, pages 2961--2969, 2017.

\bibitem{ma2018arbitrary}
Jianqi Ma, Weiyuan Shao, Hao Ye, Li~Wang, Hong Wang, Yingbin Zheng, and
  Xiangyang Xue.
\newblock Arbitrary-oriented scene text detection via rotation proposals.
\newblock {\em IEEE Transactions on Multimedia}, 20(11):3111--3122, 2018.

\bibitem{xiao2020sequential}
Shanyu Xiao, Liangrui Peng, Ruijie Yan, Keyu An, Gang Yao, and Jaesik Min.
\newblock Sequential deformation for accurate scene text detection.
\newblock In {\em European Conference on Computer Vision}, pages 108--124.
  Springer, 2020.

\bibitem{liu2019pyramid}
Jingchao Liu, Xuebo Liu, Jie Sheng, Ding Liang, Xin Li, and Qingjie Liu.
\newblock Pyramid mask text detector.
\newblock {\em arXiv preprint arXiv:1903.11800}, 2019.

\bibitem{ronneberger2015u}
Olaf Ronneberger, Philipp Fischer, and Thomas Brox.
\newblock U-net: Convolutional networks for biomedical image segmentation.
\newblock In {\em International Conference on Medical image computing and
  computer-assisted intervention}, pages 234--241. Springer, 2015.

\bibitem{paszke2019pytorch}
Adam Paszke, Sam Gross, Francisco Massa, Adam Lerer, James Bradbury, Gregory
  Chanan, Trevor Killeen, Zeming Lin, Natalia Gimelshein, Luca Antiga, et~al.
\newblock Pytorch: An imperative style, high-performance deep learning library.
\newblock {\em Advances in neural information processing systems}, 32, 2019.

\bibitem{wang2022fast}
Zi-Rui Wang and Jun Du.
\newblock Fast writer adaptation with style extractor network for handwritten
  text recognition.
\newblock {\em Neural Networks}, 147:42--52, 2022.

\bibitem{zhang2020scut}
Hesuo Zhang, Lingyu Liang, and Lianwen Jin.
\newblock Scut-hccdoc: A new benchmark dataset of handwritten chinese text in
  unconstrained camera-captured documents.
\newblock {\em Pattern Recognition}, 108:107559, 2020.

\end{thebibliography}


\begin{thebibliography}{10}

\bibitem{lecun2015deep}
Yann LeCun, Yoshua Bengio, and Geoffrey Hinton.
\newblock Deep learning.
\newblock {\em nature}, 521(7553):436--444, 2015.

\bibitem{zhong2017deeptext}
Zhuoyao Zhong, Lianwen Jin, and Shuangping Huang.
\newblock Deeptext: A new approach for text proposal generation and text
  detection in natural images.
\newblock In {\em 2017 IEEE international conference on acoustics, speech and
  signal processing (ICASSP)}, pages 1208--1212. IEEE, 2017.

\bibitem{jiang2018r}
Yingying Jiang, Xiangyu Zhu, Xiaobing Wang, Shuli Yang, Wei Li, Hua Wang, Pei
  Fu, and Zhenbo Luo.
\newblock R 2 cnn: Rotational region cnn for arbitrarily-oriented scene text
  detection.
\newblock In {\em 2018 24th International Conference on Pattern Recognition
  (ICPR)}, pages 3610--3615. IEEE, 2018.

\bibitem{ren2015faster}
Shaoqing Ren, Kaiming He, Ross Girshick, and Jian Sun.
\newblock Faster r-cnn: Towards real-time object detection with region proposal
  networks.
\newblock {\em Advances in neural information processing systems}, 28, 2015.

\bibitem{he2017mask}
Kaiming He, Georgia Gkioxari, Piotr Doll{\'a}r, and Ross Girshick.
\newblock Mask r-cnn.
\newblock In {\em Proceedings of the IEEE international conference on computer
  vision}, pages 2961--2969, 2017.

\bibitem{yan2021recognizing}
Shi Yan, Jin-Wen Wu, Fei Yin, and Cheng-Lin Liu.
\newblock Recognizing handwritten chinese texts with insertion and swapping
  using a structural attention network.
\newblock In {\em International Conference on Document Analysis and
  Recognition}, pages 557--571. Springer, 2021.

\bibitem{hu2021recyclenet}
Yiqing Hu, Yan Zheng, Xinghua Jiang, Hao Liu, Deqiang Jiang, Yinsong Liu,
  Bo~Ren, and Rongrong Ji.
\newblock Recyclenet: An overlapped text instance recovery approach.
\newblock In {\em Proceedings of the 29th ACM International Conference on
  Multimedia}, pages 1102--1110, 2021.

\bibitem{zhu2018scut}
Yuanzhi Zhu, Zecheng Xie, Lianwen Jin, Xiaoxue Chen, Yaoxiong Huang, and Ming
  Zhang.
\newblock Scut-ept: New dataset and benchmark for offline chinese text
  recognition in examination paper.
\newblock {\em IEEE Access}, 7:370--382, 2018.

\bibitem{ronneberger2015u}
Olaf Ronneberger, Philipp Fischer, and Thomas Brox.
\newblock U-net: Convolutional networks for biomedical image segmentation.
\newblock In {\em International Conference on Medical image computing and
  computer-assisted intervention}, pages 234--241. Springer, 2015.

\bibitem{paszke2019pytorch}
Adam Paszke, Sam Gross, Francisco Massa, Adam Lerer, James Bradbury, Gregory
  Chanan, Trevor Killeen, Zeming Lin, Natalia Gimelshein, Luca Antiga, et~al.
\newblock Pytorch: An imperative style, high-performance deep learning library.
\newblock {\em Advances in neural information processing systems}, 32, 2019.

\bibitem{liu2011casia}
Cheng-Lin Liu, Fei Yin, Da-Han Wang, and Qiu-Feng Wang.
\newblock Casia online and offline chinese handwriting databases.
\newblock In {\em 2011 international conference on document analysis and
  recognition}, pages 37--41. IEEE, 2011.

\bibitem{zhang2020scut}
Hesuo Zhang, Lingyu Liang, and Lianwen Jin.
\newblock Scut-hccdoc: A new benchmark dataset of handwritten chinese text in
  unconstrained camera-captured documents.
\newblock {\em Pattern Recognition}, 108:107559, 2020.

\end{thebibliography}

%\begin{thebibliography}{8}
%\bibitem{ref_article1}
%Author, F.: Article title. Journal \textbf{2}(5), 99--110 (2016)

%\bibitem{ref_lncs1}
%Author, F., Author, S.: Title of a proceedings paper. In: Editor,
%F., Editor, S. (eds.) CONFERENCE 2016, LNCS, vol. 9999, pp. 1--13.
%Springer, Heidelberg (2016). \doi{10.10007/1234567890}

%\bibitem{ref_book1}
%Author, F., Author, S., Author, T.: Book title. 2nd edn. Publisher,
%Location (1999)

%\bibitem{ref_proc1}
%Author, A.-B.: Contribution title. In: 9th International Proceedings
%on Proceedings, pp. 1--2. Publisher, Location (2010)

%\bibitem{ref_url1}
%LNCS Homepage, \url{http://www.springer.com/lncs}. Last accessed 4
%Oct 2017
%\end{thebibliography}

\end{document}